\documentclass[10pt,twocolumn,letterpaper]{article}

\usepackage{cvpr}
\usepackage{times}
\usepackage{epsfig}
\usepackage{graphicx}
\usepackage{amsmath}
\usepackage{amssymb}
\usepackage{makecell}

\usepackage{xcolor}
\usepackage{wrapfig}
\usepackage{diagbox}

\usepackage{algorithm}
\usepackage{algorithmic}

\usepackage[pagebackref=true,breaklinks=true,letterpaper=true,colorlinks,bookmarks=false]{hyperref}

 \cvprfinalcopy 

\def\cvprPaperID{9667} 

\ifcvprfinal\fi

\begin{document}


\title{Neural Networks Are More Productive Teachers Than Human Raters: \\Active Mixup for Data-Efficient Knowledge Distillation from a Blackbox Model}

    \author{Dongdong Wang $^{1}$$^{*}$  \quad Yandong Li $^{1}$$^{*}$  \quad Liqiang Wang$^{1}$   \quad Boqing Gong$^{2}$ \\
    $^{1}$University of Central Florida  \quad $^{2}$Google\\
    {\tt\small \{daniel.wang, liyandong\}@Knights.ucf.edu \quad lwang@cs.ucf.edu \quad bgong@google.com}\\
    }

\maketitle

\begin{abstract}
We study how to train a student deep neural network for visual recognition by distilling knowledge from a blackbox teacher model in a data-efficient manner. Progress on this problem can significantly reduce the dependence on large-scale  datasets for learning high-performing visual recognition models. There are two major challenges. One is that the number of queries into the teacher model should be minimized to save computational and/or financial costs. The other is that the number of images used for the knowledge distillation should be small; otherwise, it violates our expectation of reducing the dependence on large-scale datasets. To tackle these challenges, we propose an approach that blends mixup and active learning. The former effectively augments the few unlabeled images by a big pool of synthetic images sampled from the convex hull of the original images, and the latter actively chooses from the pool hard examples for the student neural network and query their labels from the teacher model. We validate our approach with extensive experiments. 
\footnote{Code and models: https://github.com/dwang181/active-mixup}.

\end{abstract}
\let\thefootnote\relax\footnote{$^{*}$ Equal contribution.}

\section{Introduction}
Data curation is one of the most important steps for learning high-performing visual recognition models. However, it is often tedious and sometimes daunting to collect large-scale relevant data that have sufficient coverage of the inference-time scenarios.  Additionally, labeling the collected data is time-consuming and costly.

\begin{figure}[t]
  \centering
  \includegraphics[width=1.0\columnwidth]{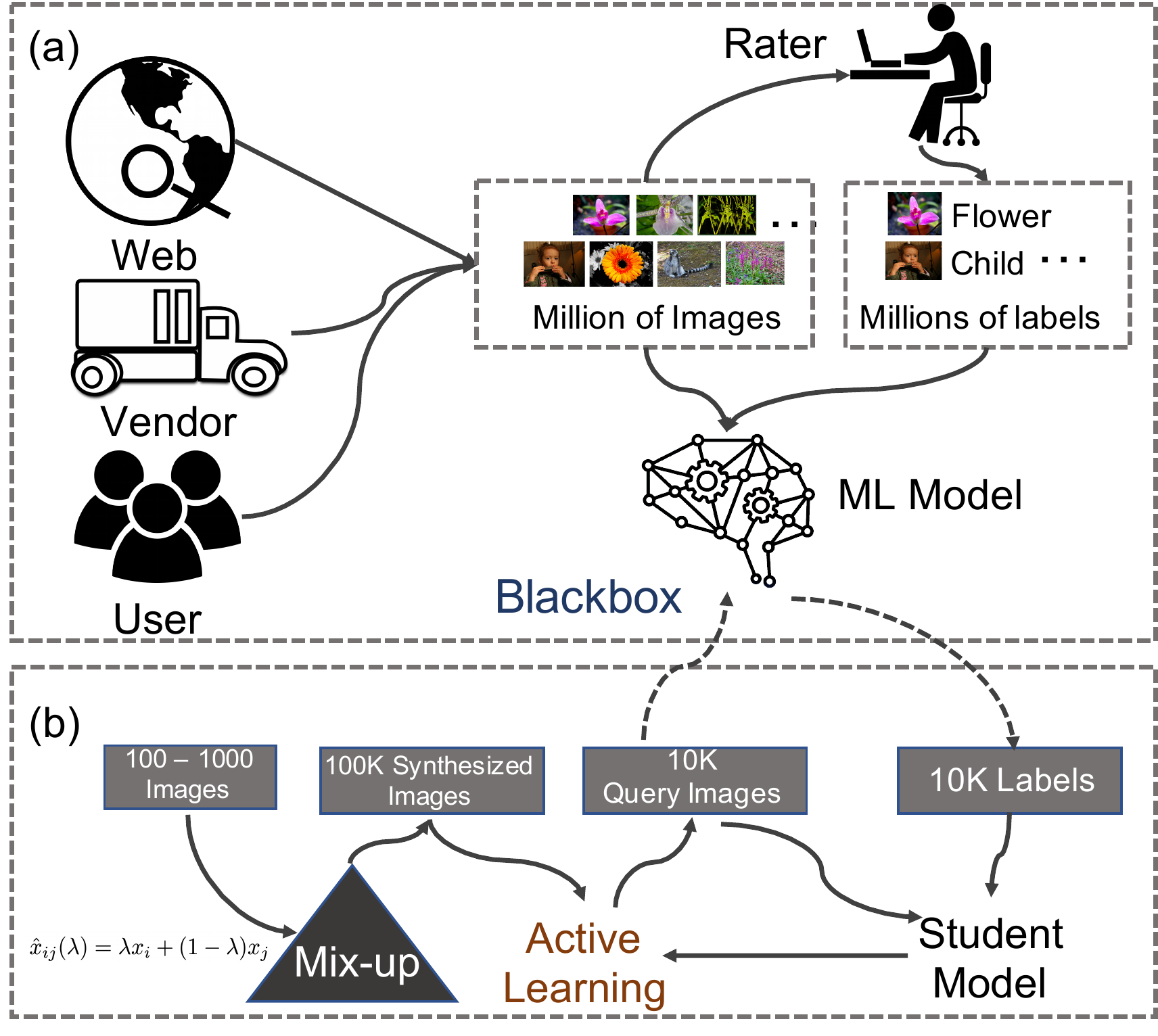}
  \caption{Data-efficient blackbox knowledge distillation. Given a blackbox teacher model and a small set of unlabeled images, we propose to employ mixup~\cite{zhang2017mixup} and active learning~\cite{lewis1994sequential} to train a high-performing student neural network in a data-efficient manner (b) so that we do not need to re-do the heavy and expensive data curation used to train the teacher model (a). }
  \label{fig:overview}
\end{figure}

Given a new task, how can we learn a high-quality machine learning model in a more data-efficient manner? We believe the answer varies depending on specific application scenarios. In this paper, we focus on the case that there exists a \emph{blackbox} teacher model whose capability covers our task of interest. Indeed, there are many high-performing generic visual recognition models available as Web-based APIs, in our smart devices, or even as an obsolete model built by ourselves some while ago. The challenge is, however, we often have limited knowledge about their specifics, e.g., not knowing the exact network architecture or weights. Moreover, it could be computationally and/or financially expensive to query the models and read out their outputs for a large-scale dataset. 

To this end, we study how to distill a \emph{blackbox} teacher model for visual recognition into a student neural network in a data-efficient manner. Our objective is three-fold. First of all, we would like the distilled student network to perform well as the teacher model as possible at the inference time. Besides, we try to minimize the number of queries to the blackbox teacher model to save  costs. Finally, we also shall use as a small number of examples as possible to save data collection efforts. It is hard to collect abundant data for rare classes or privacy-critical applications. 

We propose to blend active learning~\cite{tong2001support,lewis1994sequential} and image mixup~\cite{zhang2017mixup} to tackle the data-efficient knowledge distillation from a blackbox teacher model. The main idea is to synthesize a big pool of images from the few training examples by mixup and then use active learning to select from the pool the most helpful subset to query the teacher model. After reading out the teacher model's outputs, we simply treat them as the ``groundtruth labels'' of the query images and train the student neural network with them.

Image mixup~\cite{zhang2017mixup,guo2019mixup,berthelot2019mixmatch} was originally proposed for data augmentation to improve the generalization performance of a neural recognition network. It synthesizes a virtual image by a convex combination of two  training images. While the resultant image may become cluttered and semantically meaningless, it resides near the manifold of the natural images --- unlike white-noise images. Given 1000 images, we can construct $O(10^5)$ pairs, each of which can further generate tens to thousands of virtual images depending on the coefficients in the convex combination. We conjecture that the big pool of mixup images provides good coverage of the manifold of natural images. Hence, we expect that a student network that imitates the blackbox teacher on the mixup images can give rise to similar predictions over the test images as the teacher model does. 

Instead of querying the blackbox teacher model by all the mixup images, we resort to active learning to improve the querying efficiency. We first acquire the labels of the small number of original images from the blackbox teacher model and use them for training the student network. We then apply the \emph{student} network to all the mixup images to identify the subset with which the current student network is the most uncertain. Notably, if two mixup images are synthesized from the same pair of original images, we keep only the one with higher uncertainty. We query labels for this subset, merge it into the previously labeled data, and then re-train the student network. We iterate this procedure of subset selection, querying the blackbox teacher model, and training the student neural network multiple times until reaching a stopping criterion. 

To the best of our knowledge, we are the first to distill knowledge from a blackbox teacher model while underscoring the need for data-efficiency and query-efficiency. We empirically validate our approach by contrasting it to both vanilla and few/zero-shot knowledge distillation methods. Experiments show that, despite the blackbox teacher in our work, our approach performs on par or better than the competing methods that learn from whitebox teachers.

Note that the mixup images are often semantically meaningless, making them almost impossible for human raters to label. However, the blackbox teacher model returns predictions for them regardless, and the student network still gains from such fake image-label pairs. In this sense, we say that the blackbox teacher model is more productive than human raters in teaching the student network.

\section{Related Work}

\noindent\textbf{Knowledge Distillation.}
Knowledge distillation is proposed in \cite{hinton2015distilling} to solve model compression problems, thus relieving the burden of ensemble learning. This work suggests that class probabilities, as ``dark knowledge", are very useful to retain the performance of original network, and thus, light-weight substitute model could be trained to distill this knowledge. This approach is very useful and has been justified to solve a variety of complex application problems, such as pose estimation~\cite{saputra2019distilling,wang2019distill,nie2019dynamic}, lane detection ~\cite{hou2019learning}, real-time streaming ~\cite{mullapudi2019online}, object detection ~\cite{deng2019relation}, video representation~\cite{tavakolian2019awsd,gan2018geometry,gan2019self}, and so forth. Furthermore, this approach is able to boost the performance of deep neural network with improvement on efficiency ~\cite{phuong2019distillation} and accuracy ~\cite{kundu2019adapt}. Accordingly, lots of research is conducted to enhance its performance from the perspective of training strategy~\cite{tung2019similarity,jin2019knowledge}, distillation scheme~\cite{heo2019comprehensive, cho2019efficacy}, or network properties~\cite{peng2019correlation}
, etc.

However, there is an important issue. Traditional knowledge distillation requires lots of original training data which are very difficult to be obtained. To alleviate this data demand, few-shot knowledge distillation is proposed to retain teacher model performance with pseudo samplers which are generated in adversarial manner \cite{kimura2018few}. Another approach called data free knowledge distillation leverages extra activation records from teacher model to reconstruct original datasets, thus recovering teacher model \cite{lopes2017data}. Recently, a zero-knowledge distillation method is developed by synthesizing data with gradient information of teacher network \cite{nayak2019zero}. Nevertheless, these approaches require the gradient information of teacher network, which enables them intractable in the real world.

\paragraph{Blackbox Optimization.}
Blackbox optimization is developed based on zero knowledge in the gradient information of queried models and widely used to solve practical problems. Recently, this work is widely used in deep learning, especially model attack. A rich line of blackbox attacking approaches~\cite{chen2017zoo,ilyas2018black,salimans2017evolution,brendel2017decision,li2019nattack} are explored by accessing the input-output pairs of classifiers, most of which are focusing on attacks resulting from accessing the data. \cite{fredrikson2015model} instead investigates that the adversaries are capable of recovering sensitive data by model inversion. However, there is no work for blackbox knowledge distillation.

\paragraph{Active Learning.}
Active learning is a learning process by interaction between oracle and learner agents. This strategy is widely used to solve learning problems which exhibit costly data labelling since it could exploit existing data information to efficiently improve obtained model, thus reducing the number of queries. Lots of effective approaches are proposed to optimize this process, such as uncertainty-based  ~\cite{lewis1994sequential, yang2018benchmark,gal2017deep} and margin-based methods ~\cite{ducoffe2018adversarial, settles2008multiple}. Form the review by ~\cite{gissin2019discriminative}, uncertainty-based methods, despite simple, are able to obtain good performance.

\paragraph{Mixup.}
Zhang \textit{et al.} first proposed mixup to improve the generalization of deep neural network~\cite{zhang2017mixup}. Between-Class learning~\cite{tokozume2017learning} (BC learning) was proposed for
deep sound recognition, and then, they extended this approach to image classification~\cite{Tokozume_2018_CVPR}. Following them, Pairing Samples ~\cite{inoue2018data} was proposed as a data augmentation approach by taking an average of two images for each pixel. More recently, an approach called AutoAugment ~\cite{cubuk2018autoaugment}, explores improving data augmentation policies by automatically searching.

\section{Approach}
We present our approach to the data-efficient knowledge distillation from a blackbox teacher model in detail in this section. Given a blackbox teacher model and a small number of unlabeled images, the approach iterates over the following three steps: 1) constructing a big candidate pool of synthesized images from the small number of unlabeled images, 2) actively choosing a subset from the pool with which the current student network is the most uncertain, 3) querying the blackbox teacher model to acquire labels for this subset and to re-train the student network. 

\subsection{Constructing a Candidate Pool}
In real-world applications, data collection could consume a huge amount of time due to various reasons, such as privacy concerns, rare classes, data quality, etc. Instead of relying on a big dataset of real images, we begin with a small number of unlabeled images and use the recently proposed mixup~\cite{zhang2017mixup} to augment this initial image pool. 

Given two natural images $x_i$ and $x_j$, mixup generates multiple synthetic images by a convex combination of the two with different coefficients,
\begin{align}\label{eq:mixup}
\hat{x}_{ij}(\lambda) & = \lambda x_{i} + (1-\lambda) x_{j}, 
\end{align}
where the coefficient $\lambda\in[0,1]$. Note that this notation also includes the original unlabeled data $x_i$ and $x_j$ when $\lambda=1$ and $\lambda=0$, respectively. 

This technique comes handy and effective for our work. It can exponentially expand the size of the initial image pool. Suppose we have collected 1000 natural images, and we generate 10 mixup images for each image pair by varying the coefficient $\lambda$. We then arrive at a pool of about $10^6$ images in total. Besides, this pool of synthetic images also provides good coverage of the manifold of natural images. Indeed, this pool can be viewed as a dense sampling of the convex hull of the natural images we have collected. The test images likely fall into or close to this convex hull if the collected images are diverse and representative. Hence, we expect the student neural network to generalize well to the inference-time data by enforcing it to imitate the blackbox teacher model on the mixup images. 

\subsection{Actively Choosing a Subset to Query the Teacher Model}
Let $\{\hat{x}_{ij}(\lambda), \lambda\in[0,1], i\neq j\}$ denote the augmented pool of images. 
It is straightforward to query the teacher model to obtain the (soft) labels for these synthetic images and then train the student network with them. However, this brute-force strategy incurs high computational and financial costs. Instead, we employ active learning to reduce the cost.

We define the student neural network's confidence over an input $x$ as 
\begin{align}
    C_1(x) := \max_y P_S(y|x), \label{eq:c1}
\end{align}
where $P_S(y|x)$ is the probability of the input image $x$ belonging to the class $y$ predicted by the current student network. Intuitively, the less confidence the student network has over the input $x$, the more the student network can gain from the teacher model's label for the input.

Therefore, we could rank all the synthetic images in the candidate pool according to the student network's confidences on them, and then choose the top ones as the query subset. However, this simple strategy results in near-duplicated images, for example $\hat{x}_{ij}(\lambda=0.5)$ and $\hat{x}_{ij}(\lambda=0.55)$. We avoid this situation by choosing at most one image from any pair of images. 

In particular, instead of ranking the synthetic images, we rank image pairs in the candidate pool. We define the confidence of the student network over an image pair  $x_i$ and $x_j$ as the following,
\begin{align}\label{eq:c2}
C_2(x_i, x_j) := \min_\lambda C_1(\hat{x}_{ij}(\lambda)), \quad \lambda \in [0,1],
\end{align}
which depends on a coefficient $\lambda^*$  for the image pair. Hence, we obtain a confidence score and its corresponding coefficient for any pair of the original images. The synthetic image $\hat{x}_{ij}(\lambda^*)$ is selected into the query set if the confidence score $C_2(x_i,x_j)$ is among the lowest $k$ ones. We study the size of the query set in the experiments.

\subsection{Training the Student Network}
With the actively selected query set of images, we query the blackbox teacher model and read out its soft predictions as the labels for the images. We then merge them with the previous training set, if there is, to train the student network using a cross-entropy loss. The soft probabilistic labels returned by the teacher model give rise to slightly better results than the hard labels, so we shall use the soft labels in the experiments below.

\subsection{Overall Algorithm}

\begin{algorithm}[t]
\caption{Data-efficient blackbox knowledge distillation }
\hspace*{0.02in}
\label{algo:DEKC}

{\bf INPUT:} Pre-trained teacher model $\mathcal{M}^T$

{\bf INPUT:} A small set of unlabeled images $X = \{x_i\}_{i=1}^n$

{\bf INPUT:} Hyper-parameters (learning rate, subset size, etc.)

{\textbf{OUTPUT:}}  Student network $\mathcal{M}^S$ 
\begin{algorithmic}[1] 


\STATE{Query  $\mathcal{M}^T$ and acquire labels ${Y}_0$ for all images in $X$}

\STATE{Train an initial student network $\mathcal{M}^S_0$ with $(X,{Y}_0)$}\\

\STATE{Construct a synthetic image pool $\mathcal{P}=\{\hat{x}_{ij}(\lambda)\}$ by using the unlabeled images $X$ with eq.~(\ref{eq:mixup})}\\

\STATE{Initialize $\mathcal{P}_1^s = X, \mathcal{Y}_1 = \mathcal{Y}_0$}.\\

\FOR{ $t = 1, 2..., T$} 
    \STATE{Select a subset $\Delta\mathcal{P}_t^s$ from $\mathcal{P}$ with lowest confidence scores $\{C_2(x_i,x_j)\}$ returned by  student  $\mathcal{M}^S_{t-1}$ }\\
    \STATE{Query  $\mathcal{M}^T$, acquire labels $\Delta\mathcal{Y}_t$ for all images $\Delta\mathcal{P}_t^s$ }\\
    
    \STATE{ $\mathcal{P}_t^s \leftarrow \mathcal{P}_t^s \cup \Delta\mathcal{P}_t^s$, $\mathcal{Y}_t \leftarrow \mathcal{Y}_t \cup \Delta\mathcal{Y}_t$} \\

    \STATE{Train a new student network $\mathcal{M}^S_t$ with $(\mathcal{P}_t^s,\mathcal{Y}_t)$}\\
    
    \STATE{Update $\mathcal{P}\leftarrow \mathcal{P}$ - $\Delta\mathcal{P}_{t}^s$ } \\
\ENDFOR


\end{algorithmic}
\end{algorithm}

Algorithm~\ref{algo:DEKC} presents the overall procedure of our approach to the data-efficient blackbox knowledge distillation. Beginning with a teacher model $\mathcal{M}^T$ and a few unlabeled images $X= \{x_1,x_2,...,x_n\}$, we firstly train an initial student network $\mathcal{M}_0^S$ with $(X,{Y}_0)$, where ${Y}_0$ contains the labels for the images in $X$ and is obtained by querying the teacher model. We then  construct a big pool of synthetic images $\mathcal{P}$ with mixup~\cite{zhang2017mixup} (eq.~(\ref{eq:mixup})) to facilitate the active learning stage. We iterate the following steps until the accuracy of the student network converges. 1) Actively select a subset $\Delta\mathcal{P}^s_t$ of the synthetic images $\mathcal{P}$ with the lowest confidence scores, $C_2(x_i,x_j)$, as predicted by the current student network so that the resulting subset $\Delta\mathcal{P}^s_t$ contains hard samples for the current student network $\mathcal{M}_{t-1}^S$. 2) Acquire labels $\Delta\mathcal{Y}_t$ of the selected subset of synthetic images $\Delta\mathcal{P}^s_t$ by querying the teacher model. 3) Train a new student network $\mathcal{M}_{t}^S$ with all the labeled images thus far, $(\mathcal{P}_t^s,\mathcal{Y}_t)$. 
Note that, in Line 6 of Algorithm~\ref{algo:DEKC}, we only keep one synthetic image for any pair $(x_i,x_j)$ of the original images to reduce redundancy. 
\begin{figure*}[h]
    \centering

          \includegraphics[width=.49\columnwidth]{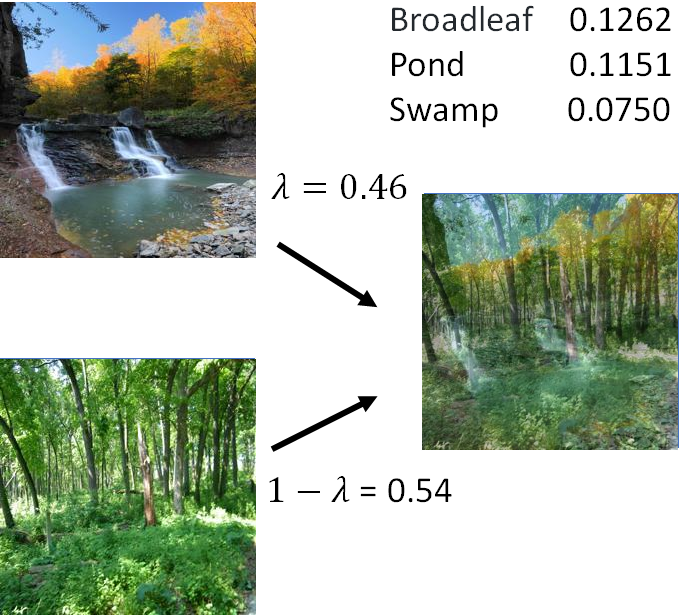}
          \hspace{0.2cm}
          \includegraphics[width=.49\columnwidth]{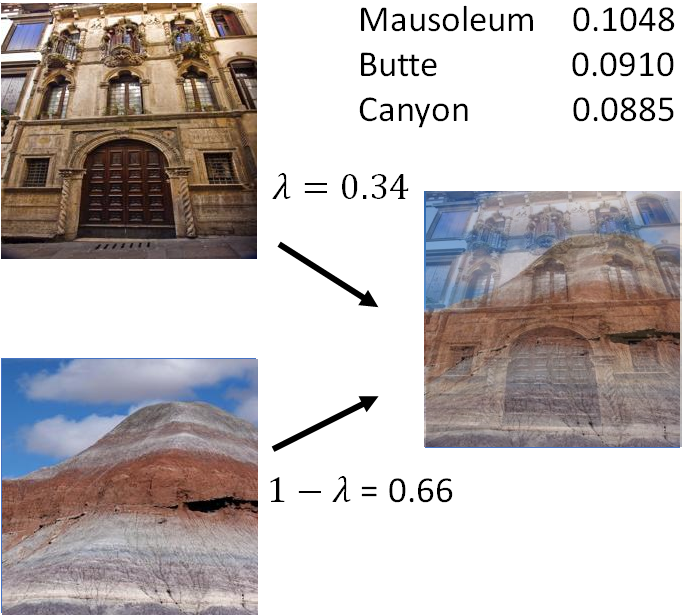}
          \hspace{0.2cm}
          \includegraphics[width=.49\columnwidth]{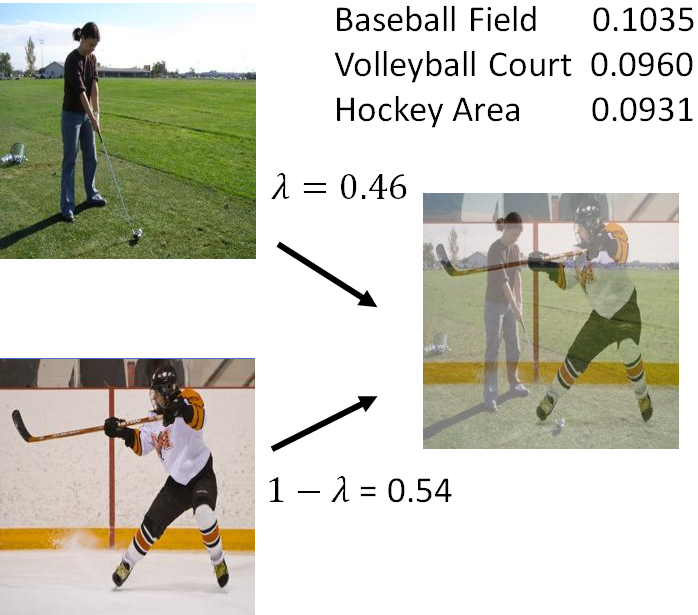}
          \hspace{0.2cm}
          \includegraphics[width=.49\columnwidth]{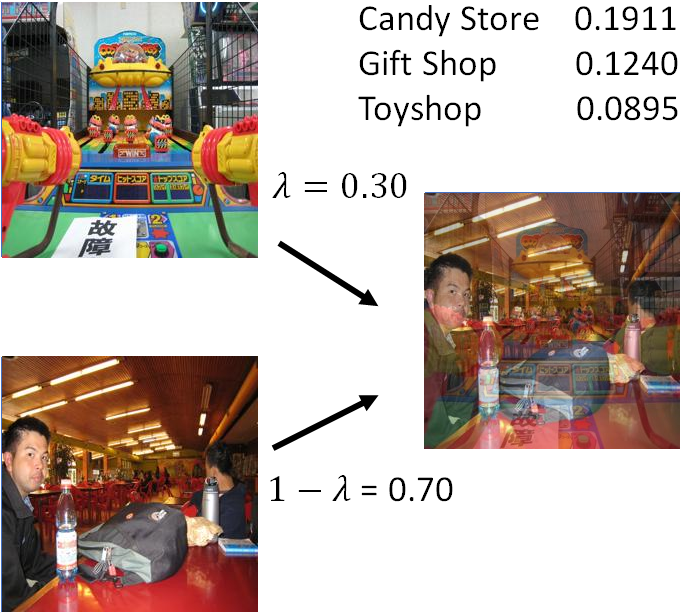}
\caption{Mixup images whose confidence scores (cf.\ eq.~(\ref{eq:c2})) are the lowest among all candidates in the third iteration. For each mixup image, we show the top three labels and probabilities returned by the blackbox teacher model.}
\label{fig:365_mixup}
\end{figure*}

\begin{figure*}[t]
  \centering
  \includegraphics[width=1\textwidth]{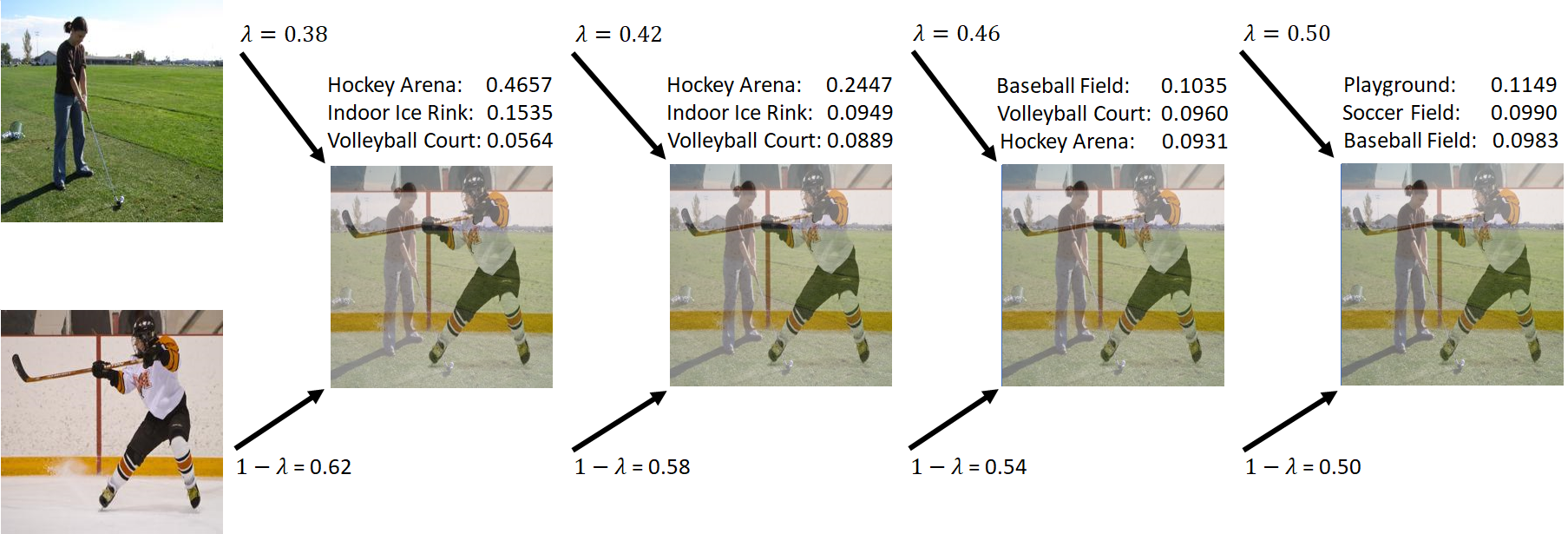}
  \caption{Different mixup images from the same pair of the original images by varying the mixup coefficient $\lambda$. We show the top three labels and probabilities predicted by the teacher model for each of them. It is interesting to see how the top-1 label changes from Hockey Arena, to Baseball Field, and to Golf Course. }
  \label{fig:365_range_lambda}
\end{figure*}

\section{Experiments} 
We design various experiments to test our approach, including both comparison experiments with state-of-the-art knowledge distillation methods and ablation studies. Additionally, we also challenge our approach when the available data is out of the distribution of the main task of interest. 
In practice, across all experiments, we select $\lambda\in\{0.3, 0.7\}$  (with an interval of 0.04) to generate synthetic images to produce more diverse mixup images.

\subsection{Comparison Experiments}
Since our main objective is to explore how to train a high-performing student neural network from a blackbox teacher model in a data-efficient manner, it is worth comparing our approach with existing knowledge distillation methods although they were developed for other setups. The comparison can help review how data-efficient our approach is given the blackbox teacher model.

\subsubsection{Experiment Setting}
\paragraph{Datasets.}
We run experiments on MNIST~\cite{lecun1998gradient}, Fashion-MNIST~\cite{xiao2017fashion}, CIFAR-10~\cite{cifar10}, and Places365-Standard~\cite{zhou2017places}, which are popular benchmark datasets for image classification. The MNIST dataset contains 60K training images and 10K testing images about ten handwritten digits. The image resolution is 28$\times$28. Fashion-MNIST is composed of 60K training and 10K testing fashion product images of the size 28$\times$28. CIFAR-10 consists of 60K (50K training images and 10K test images) 32$\times$32 RGB images in 10 classes, with 6K images per class. In addition to evaluating the proposed approach on the above described low-resolution images, we also test our approach on  Places365-Standard, which is a challenging dataset for natural scene recognition. It has 1.8M training images and 18,250 validation images in 365 classes. We use the resolution of 256$\times$256 for Places365-Standard in the following experiments.

\paragraph{Evaluation Metric.} We mainly use the classification accuracy as the evaluation metric. Additionally, we also propose a straightforward metric to measure how much ``knowledge'' the student network distills from the teacher model. This metric is computed as the ratio between the student network's classification accuracy and the teacher's accuracy, and we call it the distillation \textit{success rate}.

\begin{table*}[t]
\caption{Comparison results on Places365-Standard, CIFAR-10, MNIST, and Fashion-MNIST. The ``Teacher'' column reports the teacher model's accuracy on the test sets, ``KD Accuracy'' is the student network's test accuracy, ``Success'' stands for the distillation success rates, ``Black/White'' indicates whether or not the teacher model is blackbox, ``Queries'' lists the numbers of queries into the teacher models, and ``Unlabeled Data'' shows the numbers of original training images used in the experiments. (* results reported in the original paper)}
\label{tab:white_vs_black}
\vspace{-15pt}
\begin{center}
\begin{tabular}{l|c|c|c|c|r|r}
\hline
Task (Model) & Teacher  & KD Accuracy & Success & Black/White & Queries &  Unlabeled Data  \\
\hline
 Places365-Standard (ZSKD)~\cite{nayak2019zero}& -- &  -- &  -- &  --  &  -- & 0 \\
 \hline
Places365-Standard (FSKD~\cite{kimura2018few})&  53.69 &  38.18&  71.11 &  White  &  480,000 &  80,000 \\
 \hline
Places365-Standard (KD) &  53.69 & 49.01  & 90.35  & Black & 1,800,000 & 1,800,000 \\
\hline
Places365-Standard (Ours)
 & 53.69  & \textbf{45.71} & \textbf{85.14} & Black & 480,000 &  80,000 \\
\hline
\hline

CIFAR-10 (ZSKD)~\cite{nayak2019zero}
 & 83.03$^*$ & 69.56$^*$ & 83.78  & White & $>$2,000,000 & 0 \\
\hline
 CIFAR-10 (FSKD~\cite{kimura2018few}) & 83.07 & 40.58  &  48.85&  White  &  40,000 & 2,000 \\
\hline
CIFAR-10 (KD) & 83.07 & 80.01  & 96.31 & Black & 50,000 & 50,000\\
\hline
CIFAR-10 (Ours) & 83.07 &\textbf{74.60}  & \textbf{89.87} & Black & 40,000 & 2,000 \\
\hline
\hline

 MNIST (ZSKD)~\cite{nayak2019zero}
& 99.34$^*$ & \textbf{98.77}$^*$ & 99.42  & White & $>$1,200,000  &  0\\
\hline

MNIST (FSKD~\cite{kimura2018few})
& 99.29 & 80.43 & 81.01 &  White  & 24,000 & 2,000 \\
\hline

MNIST (KD)& 99.29 &  99.05 &  99.76&  Black & 60,000 &  60,000 \\
\hline
MNIST (Ours) & 99.29 & 98.74 & \textbf{99.45} & Black  & 24,000 & 2,000   \\
\hline
\hline

Fashion-MNIST(ZSKD)~\cite{nayak2019zero} & 90.84$^*$ & 79.62$^*$& 87.65 & White & $>$2,400,000 &  0\\
\hline
Fashion-MNIST (FSKD~\cite{kimura2018few}) & 90.80
 & 68.64 & 75.60 & White &  48,000 &  2,000 \\
\hline
Fashion-MNIST (KD)& 90.80 & 87.79  & 96.69 &  Black & 60,000 &  60,000 \\
\hline
Fashion-MNIST(Ours) & 90.80 & \textbf{80.90} & \textbf{89.10} & Black & 48,000 & 2,000 \\
\hline
\end{tabular}

\end{center}
\end{table*}

\paragraph{Blackbox Teacher Models.}  
For each task except Places365-Standard, we prepare a teacher model by following the training setting provided in~\cite{nayak2019zero}. For Places365-Standard, there is no training setting reference for the knowledge distillation research yet, so we use a pre-trained model from the dataset repository~\cite{zhou2017places} as our teacher model. On MNIST and Fashion-MNIST, we use the LeNet-5 architecture~\cite{lecun2015lenet} as the teacher model and optimize it to achieve 99.29\% and 90.80$\%$ top-1 accuracies, respectively. On CIFAR-10, we have an AlexNet~\cite{krizhevsky2012imagenet} as the teacher model and train it to obtain 83.07$\%$ top-1 accuracy.  As shown in Table~\ref{tab:white_vs_black}, the above teacher models are comparable to the teacher models in~\cite{nayak2019zero}: $83.03\%~vs.~83.07\%$ on CIFAR-10, $99.34\%~vs.~99.29\%$ on MNIST, and $90.84\%~vs.~90.87\%$ on Fashion-MNIST. For Places365-Standard, the teacher model is a ResNet-18~\cite{he2016deep} and yields 53.68$\%$ top-1 accuracy.

\paragraph{Competing Methods.}
We identify three existing relevant  methods for comparison. 
\begin{itemize}
    \item One is zero-shot knowledge distillation (ZSKD)~\cite{nayak2019zero}, which distills a student neural network with zero training example from a \textit{whitebox} teacher model. It synthesizes data by backpropagating gradients to the input through the whitebox teacher network.
    \item The second method is few-shot knowledge distillation (FSKD)~\cite{kimura2018few}, which  augments the training images by generating adversarial examples. It is the most relevant work to ours, but it depends on the computationally expensive adversarial examples~\cite{szegedy2013intriguing} and has no active learning scheme to reduce the query cost at all. The original work assumes a \textit{whitebox} teacher neural network so that it is straightforward to produce the adversarial examples, whereas there exist blackbox attack methods~\cite{li2019nattack,chen2017zoo}.
    \item The third is the vanilla knowledge distillation~\cite{hinton2015distilling}, which accesses the whole training set of the teacher model and is somehow an upper bound of our method. 
\end{itemize}
 

\subsubsection{Quantitative Results}
Table~\ref{tab:white_vs_black} shows the comparison results. For simplicity, we run the active learning stage for only one step (i.e., $T=1$ in Algorithm 1). Section~\ref{sec:4.2} presents the results of running it for multiple steps.

\paragraph{Accuracy.} Our approach significantly outperforms FSKD over all the datasets. On CIFAR-10, MNIST, and Fashion-MNIST, ours yields 41\%, $18\%$, and $14\%$ success rate improvements over FSKD, respectively. On Places365-Standard, whose images are high-resolution about natural scenes, we also outperform FSKD by 14\% success rate. Compared to ZSKD, which relies on a whitebox teacher network, our approach also shows higher accuracies and success rates except on MNIST. We were not able to reproduce ZSKD on Places365-Standard because its images are all high-resolution, making it computationally infeasible to generate a large number of gradient-based inputs. Similarly, the advantage of ours over ZSKD is larger on CIFAR-10 than other MNIST or Fashion-MNIST, probably because the CIFAR-10 images have a higher resolution. In contrast, the computation cost of our active mixup approach does not depend on the input resolution. Overall, the results indicate that active mixup has a higher potential to solve the larger-scale knowledge distillation in a data-efficient manner.

\paragraph{Queries.} Our approach saves orders of queries into the teacher model  compared to ZSKD. For example, we only query the blackbox teacher model up to 40K times for CIFAR-10. In contrast, ZSKD requires more than 2M queries and yet yields lower accuracy than ours. The big difference is not surprising because the gradient-based inputs in ZSKD are less natural than or representative of the test images than our mixup images. Besides, ZSKD incurs additional queries into the whitebox teacher model every time it produces an input.

\subsubsection{Qualitative Intermediate Results}
We show some mixup images in Figures~\ref{fig:365_mixup} and~\ref{fig:365_range_lambda}. These images are selected from the candidate pool constructed using the natural images in the Places365-Standard training set. Figure~\ref{fig:365_mixup} shows some mixup images with low confidence scores. They can potentially benefit the student network more than the other candidate images if we use them to query the teacher model. Figure~\ref{fig:365_range_lambda} demonstrates some mixup images synthesized from the same pair of natural images by varying the mixup coefficient $\lambda$. It is interesting to see that the mix of ``Hockey Arena'' and ``Golf Course'' leads to a ``Baseball Field'' at $\lambda=0.46$ predicted by the blackbox teacher model. This indicates that our active mixup approach can effectively augment the originally small training set by not only bringing in new synthetic images but also comprehensive coverage of classes.

\subsection{Ablation Study}\label{sec:4.2}

We select CIFAR-10 and Places365-Standard to study our approach in detail since they represent the small-scale and large-scale settings, respectively. For CIFAR-10, we switch to VGG-16~\cite{simonyan2014very} as the blackbox teacher model, which gives rise to 93.31$\%$ top-1 accuracy. 

\subsubsection{Data-Efficiency and Query-Efficiency}
We investigate how the results of our active mixup approach change as we vary the total number of unlabeled real images (data-efficiency) and the number of synthetic images selected by the active learning scheme (query-efficiency). Here we run only one step of the active learning stage ($T=1$ in Algorithm 1) to save computation cost.  Tables~\ref{tab:image_vs_mix_cifar10} and \ref{tab:image_vs_mix_place365} show the results on CIFAR-10 and Places365-Standard, respectively. Each entry in the tables is a classification accuracy on the test set, and it is obtained by a student network which we distill by using the corresponding number of unlabeled real images (Real images) and the number of selected synthetic images (Selected Syn.).

\begin{table}[h]
\caption{ Classification accuracy on CIFAR-10 with different numbers of real images and selected synthetic images.}
\label{tab:image_vs_mix_cifar10}
\vspace{-15pt}
\begin{center}
\resizebox{\columnwidth}{!}{%
\begin{tabular}{c|c|c|c|c|c|c}
\hline
\diagbox[width=8em]{Selected Syn.}{Real images} & 0.5K & 1K  & 2K & 4K&  8K &16K  \\
\hline
0 & 44.72& 56.87 &68.09 & 76.59& 83.61 & 86.89 \\
\hline
5K & 66.97 & 71.67 & 77.76& 81.76& 85.76 &87.05\\
\hline
10K & 73.60 & 77.27 & 81.27 &83.27 & 86.56 & 88.79\\
\hline
20K &77.44 & 81.18  & 84.19 & 86.29 & 88.07 & 89.01\\
\hline
40K & 82.28&  84.25 & 86.06 & 87.71 &  89.00 &  90.49\\
\hline
80K & 85.18 & 86.53& 87.89 & 88.71& 89.61 & 90.96 \\
\hline
160K & 86.56 & 88.94 & 89.42 & 90.26 & 90.87& 91.51\\
\hline

\end{tabular}
}
\end{center}
\end{table}

\begin{table}[h]
\caption{ Classification accuracy on Places365-Standard with different numbers of real images and selected synthetic images.}
\label{tab:image_vs_mix_place365}
\vspace{-5pt}
\begin{center}
{
\begin{tabular}{c|c|c|c}
\hline
\diagbox[width=9em]{Selected Syn.}{Real images} & 20K & 40K & 80K \\
\hline
100K &40.72 & 41.95 &  43.52\\
\hline
200K & 41.15& 42.86&  44.77\\
\hline
400K &41.94 & 43.42 & 45.71\\
\hline
\end{tabular}
}
\end{center}
\end{table}

We can see that the more synthetic images we select by their confidence scores (cf.\ eq.~(\ref{eq:c2})), the higher-quality the distilled student network is. It indicates that the mixup images can effectively boost the performance of our method. Meanwhile, the higher the number of unlabeled real images we have, the higher the distillation success rate we can achieve. What's more interesting is that, when the number of synthetic images is high (e.g., 160K), the gain is diminishing as we increase the number of real images. Hence, depending on the application scenarios, we have the flexibility to trade-off the real images and synthetic images for achieving a certain distillation success rate.


We can take a closer look at Tables~\ref{tab:image_vs_mix_cifar10} and \ref{tab:image_vs_mix_place365} to obtain an understanding about the ``market values'' of the selected synthetic images. In Table~\ref{tab:image_vs_mix_cifar10}, 10K selected synthetic images and 8K unlabeled real images yield 86.56$\%$ accuracy;  20K synthetic images and 4K real images lead to 86.29$\%$ accuracy; and 40K synthetic images with 2K real examples give rise to 86.06$\%$ accuracy. The accuracies are about the same. There is a similar trend along the off-diagonal entries in Table~\ref{tab:image_vs_mix_place365}, implying that if we reduce the number of real images by half, we can complement it by doubling the size of synthetic images to maintain about the same distillation success rate.

\subsubsection{Active Mixup vs.\ Random Search}

We design another experiment to compare active mixup with the random search to understand the effectiveness of our active learning scheme. We keep 500 real images for CIFAR-10 and 20K for Places365-Standard. We then use them to construct 100K and 300K synthetic images, respectively. For a fair comparison, we let random search and active mixup share the same sets of natural images. Since our active learning scheme avoids selecting redundant images by using the improved confidence score in eq.~(\ref{eq:c2}), we also equip the random search such capability by using a single mixup coefficient of $\lambda=0.5$ to construct the synthetic images. This guarantees that, like our approach, no two synthetic images selected by the random search are from the same pair of real images.

\begin{figure}[h]
  \centering
  \includegraphics[width=0.5\textwidth]{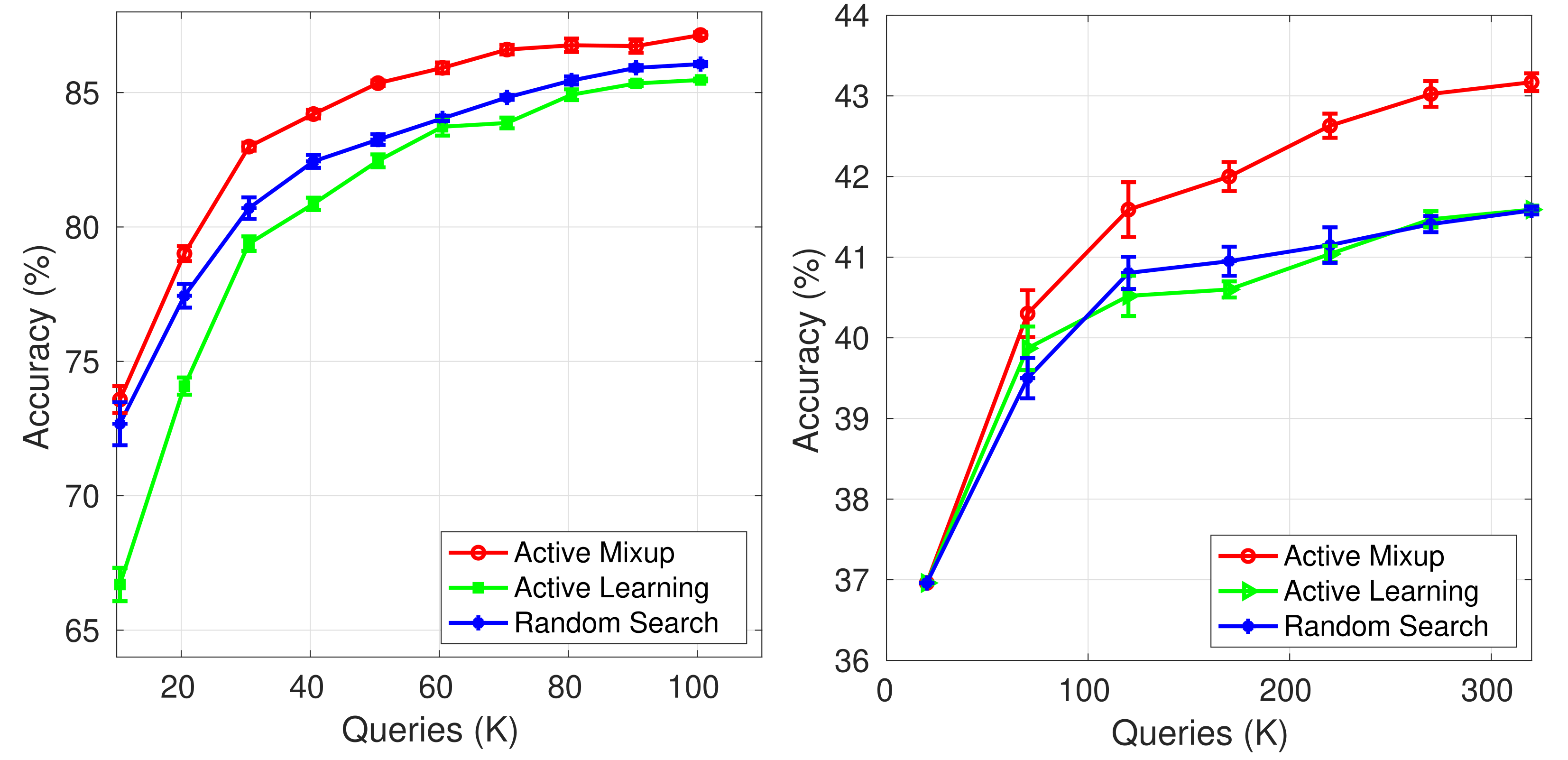}
  \caption{Test accuracy of  student networks vs.\ number of queries into the blackbox teacher model on CIFAR-10 (left) and Places365-Standard (right). We use 500 and 20K natural images for the two datasets, respectively. The plot for CIFAR-10 starts from first active learning stage ($t=1$ in Algorithm 1) and the one for Places365 starts from the initial student network training by natural images. The initial student network for CIFAR-10 trained by using natural images only yields 43.67\% accuracy.}
  \label{fig: active_mixup}
\end{figure}

Figure~\ref{fig: active_mixup} shows the comparison results of our active mixup and the random search. 
On CIFAR-10, we select 10K synthetic images every time and run the active learning stage for 10 steps ($T=10$ in Algorithm 1). On Places365-Standard, we run it for six steps and choose 50K synthetic images per step. We can see that active mixup significantly outperforms random search over the whole course of knowledge distillation, verifying its effectiveness on improving the query-efficiency. More concretely, 80K actively selected synthetic images yield $86.76\%$ accuracy, which is about the same as what 160K randomly selected synthetic images can achieve on CIFAR-10. Similarly, 40K synthetic images by active mixup lead to $84.2\%$ accuracy, on par with the $85.18\%$ accuracy by 80K randomly chosen synthetic images.



\subsubsection{Active Mixup vs.\ Vanilla Active Learning}
Our active learning scheme (eq.~(\ref{eq:c2})) improves upon the vanilla score-based active learning (eq.~(\ref{eq:c1})) by selecting only one synthetic image at most from any pair of real images. This change is necessary because two nearly duplicated synthetic images could both have very low scores according to eq.~(\ref{eq:c1}). 


To quantitatively compare the two active learning methods, we run another experiment by replacing our active learning scheme with the vanilla version. The candidate pool is the same as ours, i.e., mixup images generated by varying $\lambda\in\{0.3, 0.7\}$ with an interval of 0.04. Figure~\ref{fig: active_mixup} shows the results on both CIFAR-10 and Place365-Standard.


Generally, the vanilla active learning yields lower accuracy than our active mixup and the random search. This shows that the vanilla score-based active learning even fails to improve upon random search because it selects nearly duplicated synthetic images to query the teacher model. In contrast, our active mixup consistently performs the better than the vanilla active learning and random search. The prominent gap justifies that the constraint by $C_2$ in eq.~(\ref{eq:c2}) is crucial in our approach.




\subsection{Active Mixup with Out-of-Domain Data for Blackbox Knowledge Distillation}
In real-world applications, it may be hard to collect real training images for some tasks, e.g., due to privacy concerns. Under such scenarios, we have to use out-of-domain data to distill the student neural network. Hence, we further challenge our approach by revealing some images that are out of the domain of the training images of the blackbox teacher model. 

We conduct this experiment on CIFAR-10 by providing our approach some training images in CIFAR-100~\cite{krizhevsky2009learning}. To reduce information leak, we exclude the images that belong to the  CIFAR-10 classes and keep 2K images to construct the candidate pool. 
Equipped with these synthetic images, we run active mixup to distill student neural networks from a blackbox teacher model for CIFAR-10. The teacher model is VGG-16, which yields 93.31\% accuracy on the CIFAR-10 test set. 


\begin{table}[h]
\caption{CIFAR-10 classification accuracy by the student neural networks which are distilled by using out-of-domain data. }
\label{tab:cifar100_KD_SI}

\begin{center}
{
\begin{tabular}{c|c|c|c|c}
\hline
Selected Syn. & 10K & 20K & 40K & 80K \\
\hline
Accuracy ($\%$) & 64.10 & 71.39 & 77.89 & 83.03\\
\hline
\end{tabular}
}
\end{center}

\end{table}

Table~\ref{tab:cifar100_KD_SI} shows the results of different numbers of selected synthetic images. We still run only one iteration of the active learning to save computation costs. The best distillation performance is 83$\%$ top-1 accuracy and success rate is 88.9$\%$. Comparing the result to Table~\ref{tab:image_vs_mix_cifar10}, especially the entry (87.89\%) of 80K selected synthetic images and 2K real images, we can see that our approach  leads to about the same performance by using the out-of-domain data as the in-domain data.  

\begin{table}[h]
\caption{CIFAR-10 classification accuracy by the student neural networks which are distilled by using out-of-domain data. We set the number of selected synthetic images to 40K and vary the numbers of real images. }
\label{tab:cifar100_KD_UD}

\begin{center}
{
\begin{tabular}{c|c|c|c|c}
\hline
Real images & 500 & 1000 & 1500 & 2000 \\
\hline
Accuracy ($\%$) & 70.21 & 74.60 & 75.54 & 77.89 \\
\hline
\end{tabular}
}
\end{center}
\end{table}

To better understand how different factors influence the distillation performance, we also decouple the number of available real images from the number of selected synthetic images in Table~\ref{tab:cifar100_KD_UD}. We fix the number of selected synthetic images to 40K and vary the numbers of real images. Not surprisingly, the more real images there are, the higher distillation accuracy the active mixup achieves. Furthermore, the number of synthetic images still plays a prominent role in distillation accuracy, according to Table~\ref{tab:cifar100_KD_SI}. Without the original training data, mixup augmentation is probably more critical to enhancing the distillation performance than otherwise.

\section{Discussion and Conclusion}
In this paper, we formalize a novel problem, knowledge distillation from a blackbox teacher model in a data-efficient manner, which we think is more realistic than previous knowledge distillation setups. There are two key challenges to this problem. One is that the available examples are insufficient to represent the vast variation in the original training set of the teacher model. The other is that the blackbox teacher model often implies that it is financially and computationally expensive to query. 

To deal with the two challenges, we propose an approach combining mixup and active learning. Although neither of them is new by itself, combining them is probably the most organic solution to our problem setup for the following reasons. First of all, we would like to augment the few available examples. Unlike conventional data augmentations (e.g., cropping, adding noise), which only probe the regions around the available examples, mixup provides a continuous interpolation between any pairwise examples. As a result, mixup allows the student model to probe diverse regions of the input space. We then employ active learning  to reduce the query transactions to the teacher model. Extensive experiments verify the effectiveness of our approach to the data-efficient blackbox knowledge distillation.

\section{Acknowledgements} 
This work was supported in part by NSF-1741431 and NSF-1836881.

\end{document}